\documentclass[a4paper]{llncs}

\usepackage{geometry}
\usepackage{amssymb}
\usepackage{graphicx}
\usepackage{amsmath}
\usepackage{caption}
\usepackage{subfig}
\usepackage{csvsimple}
\usepackage{multirow}
\usepackage{array}
\usepackage{booktabs}

\usepackage{url}

\begin{document}


\title{Towards the identification of Parkinson's Disease using only T1 MR Images}


%
%
\author{Sara Soltaninejad\and Irene Cheng\and Anup Basu}


\institute{Dept. of Computing Science, University of Alberta, Canada}

%
%

\maketitle

\begin{abstract}
Parkinson\textsc{\char13}s Disease (PD) is one of the most common types of neurological diseases caused by progressive degeneration of dopaminergic neurons in the brain. Even though there is no fixed cure for this neurodegenerative disease, earlier diagnosis followed by earlier treatment can help patients have a better quality of life. Magnetic Resonance Imaging (MRI) has been one of the most popular diagnostic tool in recent years because it avoids harmful radiations. In this paper, we investigate the plausibility of using MRIs for automatically diagnosing PD. Our proposed method has three main steps : 1) Preprocessing, 2) Feature Extraction, and 3) Classification. The FreeSurfer library is used for the first and the second steps. For classification, three main types of classifiers, including Logistic Regression (LR), Random Forest (RF) and Support Vector Machine (SVM), are applied and their classification ability is compared. The Parkinson’s
Progression Markers Initiative (PPMI) data set is used to evaluate the proposed method. The proposed system prove to be promising in assisting the diagnosis of PD.
\end{abstract}

\section{Introduction}
Parkinson's Disease (PD) is the second most important neurodegenerative disease after Alzheimer's Disease (AD) that affects middle aged and elderly people. The statistical information presented by Parkinson’s News Today \cite{1} shows that an estimated seven to ten million people worldwide have Parkinson’s disease. PD causes a progressive loss of dopamine generating neurons in the brain resulting in two types of symptoms, including motor and non-motor. The motor symptoms are bradykinesia, muscles rigidity, tremor and abnormal gait \cite{2}, whereas non-motor symptoms include mental disorders, sleep problems, and sensory disturbance \cite{3}.
Even though there are some medical methods of diagnosing and determining the progress of PD, the results of these experiments are subjective and depend on the clinicians' expertise. On the other hand, clinicians are expensive and the process is time consuming for patients \cite{4}. Neuroimaging techniques have significantly improved the diagnosis of neurodegenerative  diseases. There are different types of neuro imaging techniques of which Magnetic Resonance Imaging (MRI) is one of the most popular because it is a cheap and non-invasive method. People with PD exhibit their symptoms when they lose almost $80\%$ of their brain dopamine \cite{5}. All of these facts prove the urgent need to have a Computer Aided Diagnosis (CAD) system for an automatic detection of this type of disease.
In recent years machine learning has shown remarkable results in the medical image analysis field. The proposed CAD system in neuro disease diagnosis uses different types of imaging data, including Single-Photon Emission Computed Tomography (SPECT) (Prashanth et al. \cite{6}), diffusion tension imaging (DTI), Positron Emission Tomography (PET)(Loane and Politis \cite{7}) and MRI. In this study, the goal is to utilize a structural MRI (sMRI) for developing an automated CAD to early diagnose of PD. Focke et al. \cite{8} proposed a method for PD classification using MR Images. The proposed method in \cite{8} used Gray Matter (GM) and White Matter (WM) individually with an SVM classifier. Voxel-based morphometry (VBM) has been used for preprocessing and feature extraction. The reported results show poor performance (39.53\%) for GM and 41.86\% for WM. Babu et al-\cite{9}  proposed a CAD system for diagnosing PD. Their method include three general steps: feature extraction, feature selection, and classification. In the first part, the VBM is used over GM to construct feature data. For the feature selection, recursive feature elimination (RFE) was used to select the most discriminative features. In the last step,  projection based learning and meta-cognitive radial basis function was used for classification, which resulted in 87.21\% accuracy. The potential biomarker for PD is identified as the superior temporal gyrus. The limitation in this work is that VBM is univariate and RFE is computationally expensive.
Salvatore et al. \cite{9}, proposed a method that used PCA for feature extraction. The PCA was applied to normalized skull stripped MRI data. Then, SVM was used as the classifier, resulting in 85.8\% accuracy.
Rana et al. \cite{10} extracted features over the three main tissues of the brain consisting of WM, GM and CSF. Then, they used t-test for feature selection and in the next step, SVM for classification. This resulted in 86.67\% accuracy for GM and WM and 83.33\% accuracy for CSF. In their other work \cite{11},  graph-theory based spectral feature selection method was applied to select a set of discriminating features from the whole brain volume. A decision model was built using SVM as a classifier with a leave-one-out cross-validation scheme, giving 86.67\% accuracy. The proposed method in \cite{4} was not focused on just individual tissues (GM,WM and CSF); rather, it considered the relationship between these areas because the morphometric change in one tissue might affect other tissues. $3D$ LBP was used as a feature extraction tool that could produce structural and statistical information. After that, minimum redundancy and maximum relevance with t-test are used as a feature selection methods to get the most discriminative and non-redundant features. In the end, SVM is used for classification giving 89.67\% accuracy.
In \cite{13}, the low level features (GM, cortical volume, etc.) and the high level features (region of interest (ROI) connectivity) are combined to perform a multilevel ROI feature extraction. Then, filter and wrapper feature selection method is followed up with multi kernel SVM to achieve 85.78\% accuracy for differentiation of PD and healthy control (HC) data.
Adeli et al \cite{14} propose a method for early diagnosis of PD based on the joint feature-sample selection (JFSS) procedure,  which not only selects the best subset of most discriminative features, but also it is choosing the best sample to build a classification model. They have utilized the robust regression method and further develop a robust classification model for designing the CAD for PD diagnosis. They have used MRI and SPECT images for evaluation on both synthetic and publicly available PD datasets which is shown high accuracy classification. 

In this paper, a CAD is presented for diagnosing of PD by using MR T1 Images. The general steps of the proposed method is shown in Fig.\ref{fig:framework} including preprocessing, feature extraction and classification. 

The remaining sections of this paper are structured as follows: Section 2 and 3 presents materials and methods, which provides details of the dataset, preprocessing and the proposed method for PD classification. The experimental results and discussion are provided in Section 4. Section 5 shows the conclusion.

\section{Dataset}
The data used in the preparation of this article is the T1-weighted brain MR images obtained from the PPMI database (\url{www.ppmi-info.org/data}). PPMI is a large-scale, international public study to identify PD progression biomarkers \cite{15}. The data that is used in our study contains the original T1 MR image of $598$ samples with $411$ Parkinson disease (PD) and $187$ healthy control (HC). Furthermore, the data also includes demographic or clinical information on the age and sex of the subjects. The summary of the data base is presented in Table \ref{tab:ppmi}. Based on the demographic information in this table, the balance of dataset is presented for the two type of classes which are PD and HC.
\begin{table}[h]\small
  \centering
  \caption{Demographics of the PPMI}
  \begin{tabular}{|>{\bfseries}c|*{7}{c|}}\hline
    \multirow{2}{*}{\bfseries Data Type}
    & \multicolumn{2}{c|}{\bfseries Class}
    &\multicolumn{2}{c|}{\bfseries Sex}
    & \multicolumn{3}{c|}{\bfseries Age} \\\cline{2-8}
    & \textbf{PD} & \textbf{HC} & \textbf{F}&\textbf{M} & \textbf{(25-50)}&\textbf{(50-76)} &\textbf{(75-100)}\\ \hline
           Number of Subjects   &411 & 187 & 217 & 381 & 81&472&45      \\ \hline
  \end{tabular}
  \label{tab:ppmi}
\end{table}


\section{Proposed Method}
\begin{figure}[h!]
\centering
\includegraphics[scale=0.37]{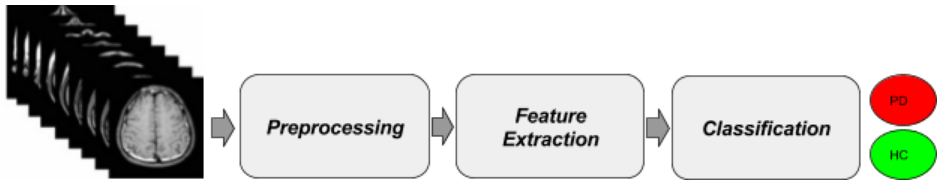}
\caption{The general framework of the proposed methods.}
\label{fig:framework}
\end{figure}

The framework of our proposed method presented in Fig.\ref{fig:framework} that includes 3 general steps: 1- Preprocessing; 2- Feature Extraction; and 3- Classification. The goals of CAD system are:
\begin{enumerate}
\item Extract the volume based features from the MR T1 images using FreeSurfer. 
\item Comparing the capability of different type of classifier for diagnosis PD
\end{enumerate}

\subsection{Preprocessing}
\begin{figure}
\centering
\includegraphics[scale=0.4]{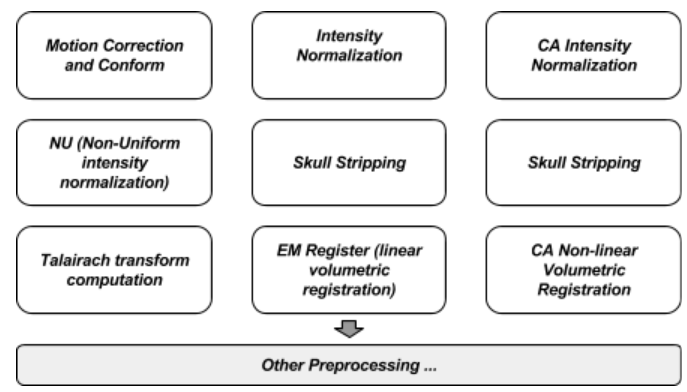}
\caption{Preprocessing steps.}
\label{fig:prepsteps}
\end{figure}

Preprocessing is an essential step in designing the CAD system providing an informative data for the next steps. In this paper, we used several preprocessing steps to compute the volumetric information of the MRI subject’s. The FreeSurfer image analysis suite is used to perform preprocessing of the 3D MRI data. FreeSurfer is a software packageto analyze and visualize structural and functional neuroimaging data from cross-sectional or longitudinal studies \cite{16}. he FreeSurfer library is proposed to do cortical reconstruction and subcortical volumetric segmentation and preprocessing including the removal of non-brain tissue (skull, eyeballs and skin), using an automated algorithm with the ability to successfully segment the whole brain without any user intervention \cite{f}. FreeSurfer is the software for structural MRI analysis for the Human Connectome Project which the documentation can be downloaded on-line (http://surfer.nmr.mgh.harvard.edu/). In total 31 preprocessing steps has been done by using FreeSurfer which some of them are shown in Fig.\ref{fig:prepsteps}.

There are two types of failures occurring in the preprocessing step: hard failures and soft failures. Hard failures apply to the subjects for whom preprocessing has not been successful; soft failures apply to the subjects who have been preprocessed but there are some problems in the preprocessing results which affect the results of the next analysis. Out of $568$ subjects MRIs, $507$ images were successfully preprocessed. Other images were excluded from
the dataset due to poor quality of the original images or unknown CDR labels.

\subsection{Feature Extraction}
After preprocessing using FreeSurfer, a list of volume based features is extracted from different regions of the brain. These features were captured from the regions segmented through brain parcellation using FreeSurfer. Some of the features collected in the left and right hemispheres of the brain are listed below:
\begin{enumerate}
	\item Left and right lateral ventricle
	\item Left and right cerebellum white matter
	\item Cerebrospinal fluid (CSF)
	\item Left and right hippocampus
	\item Left and right hemisphere cortex
	\item Estimated total intra cranial (eTIV)
	\item Left and right hemisphere surface holes
\end{enumerate}
The extracted feature data is based on Equation \ref{eq:fd}.
\begin{equation}
FeatureData =
\begin{bmatrix}
    f_{11}       & f_{12} & f_{13} & \dots & f_{1n} \\
    f_{21}       & x_{22} & x_{23} & \dots & f_{2n} \\
    \qquad \ldots\\
    f_{s1}       & f_{s2} & f_{s3} & \dots & f_{sn}
\end{bmatrix}
\label{eq:fd}
\end{equation}
where $s$ is the number of subjects and $n$ is the number of extracted features for that subject. In this study, $n$ is $507$ and $m$ is $139$.

Furthermore, there are two other types of features provided by the PPMI dataset : each subject's age and sex. Thus, these two pieces of biographical information could be added to the extracted feature from FreeSurfer.
\subsection{Classification}
In this part, our goal is to use the extracted volume based features to classify the MRI data into two classes of PD and HC. In our study, three types of supervised classification algorithms are used. Next, each classification method is described:
\begin{itemize}
\item \textbf{Logistic Regression (LR):}\\
Logistic regression (LR) is a statistical technique which is used in machine learning for binary classification. LR belongs to the group of MaxEnt classifiers known as the exponential or log-linear classifiers \cite{LR}. LR belongs to the family of classifiers known as the exponential or log-linear classifiers \cite{LR}. It is following three general steps including: Extraction of weights features from the input, Taking log , and linearly combination of them\cite{17}. 
\item \textbf{Random Forest (RF):}\\
Random forests (RF) is an ensemble learning method for classification, regression and other tasks. This method is presented by Breiman \cite{18}, which creates a set of decision trees (weak classifier) from randomly selected subset of training data. It then aggregates the votes from different decision trees to decide the final class of the test object. In the current stage of this research, we tested how accurate decisions can be made
by RF with the data coming from a the PD's MRI volumes.
\item \textbf{Support Vector machine (SVM):} \\
Support vector machine (SVM) \cite{19} is a well-known supervised machine learning algorithm for classification and regression. It performs classification tasks by making optimal hyperplanes in a multidimensional space that distinguish different class of data. This classification method is more popular because its easier to use, has higher generalization performance and little tunning comparing to other classifier. In our case, the kernel SVM is used. 

\end{itemize}

There is a set of parameters for each classifier that needs to be tuned in order to have a fair comparison.

\section{Results and Discussion}
\begin{figure}[!h]\small
  \centering
  \subfloat[Original MR image.]{\includegraphics[scale=0.25]{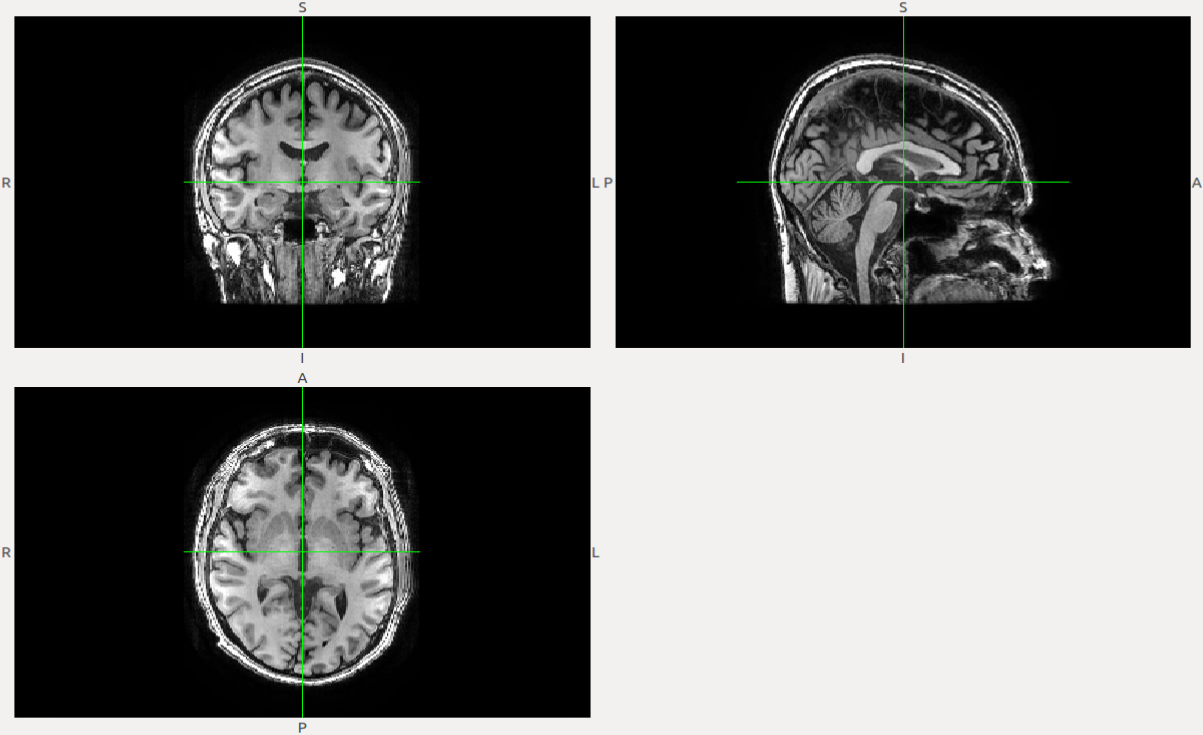}\label{fig:f1}}
  \hfill
  \subfloat[Preprocessed MR image.]{\includegraphics[scale=0.25]{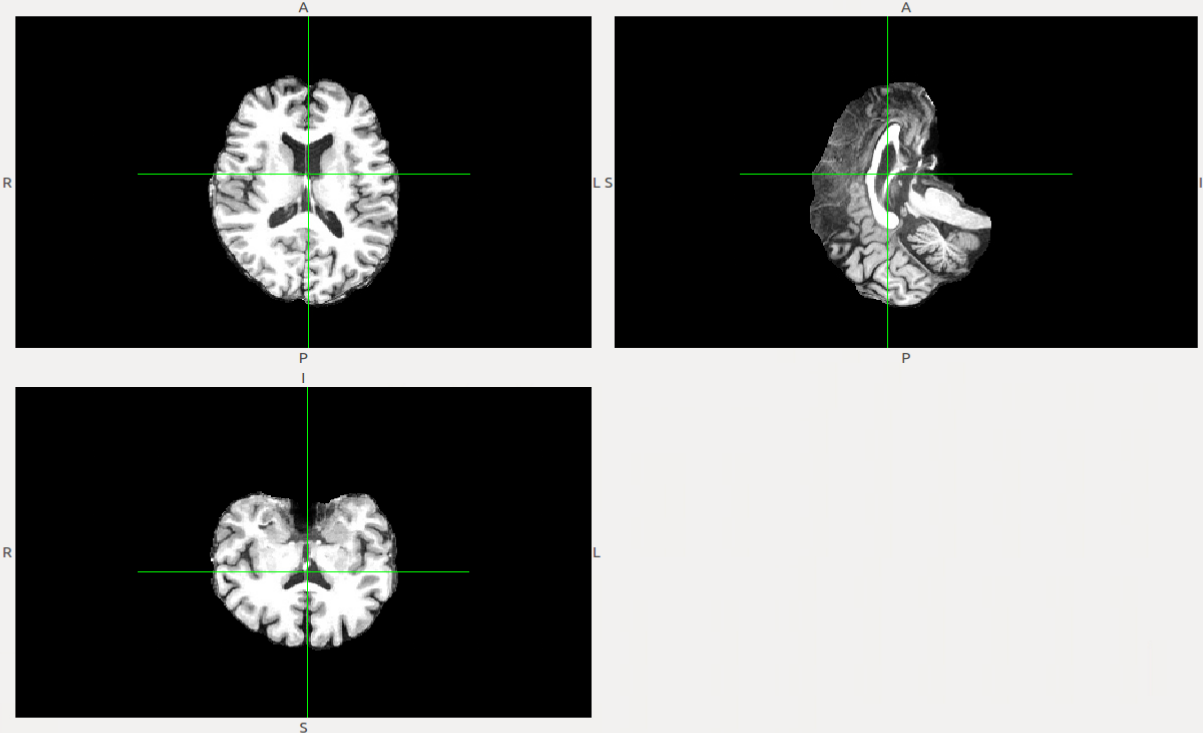}\label{fig:f2}}
  \caption{Preprocessing results for one of the subjects.}
  \label{fig:prepres}
\end{figure}

In this section, we present the experimental results of the  different steps of the proposed CAD system to diagnose PD is presented. First, using FreeSurfer, the preprocessing step prepares the MRI data for the next steps. Fig.\ref{fig:prepres} shows the MRI for subject $3102$ and the resulting image after preprocessing.

\begin{figure}[h]\small
\centering
\includegraphics[scale=0.4]{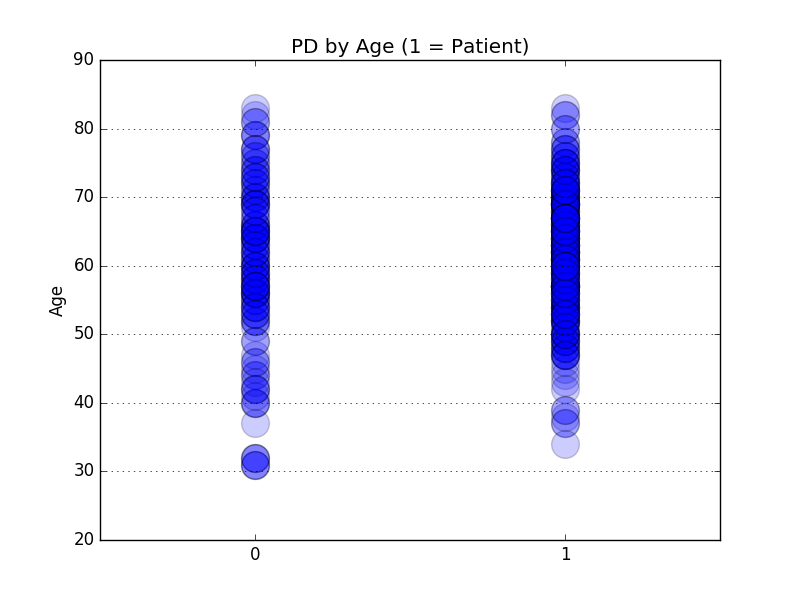}
\caption{Distribution of Data in terms of Age feature.}
\label{fig:agelabel}
\end{figure}

\begin{figure}[h]\small
  \centering
  \subfloat[]{\includegraphics[width=0.5\textwidth]{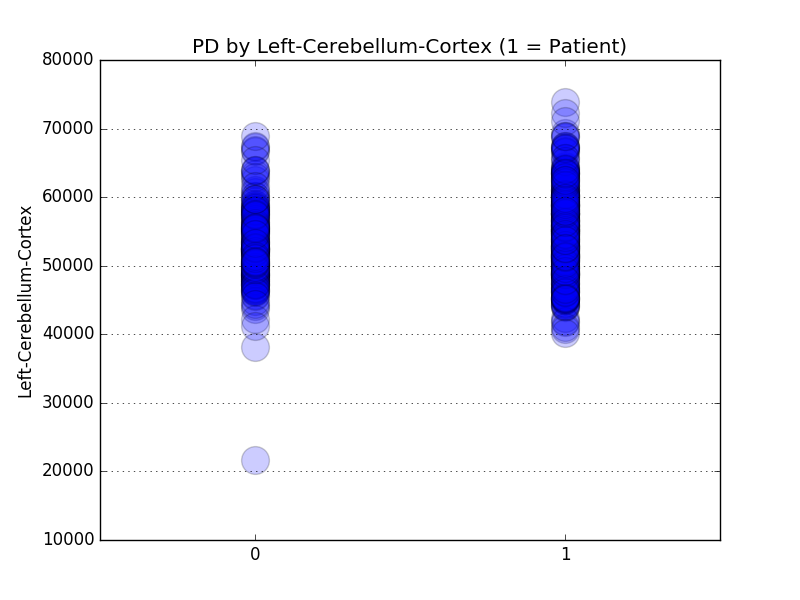}\label{fig:f1}}
  \hfill
  \subfloat[]{\includegraphics[width=0.5\textwidth]{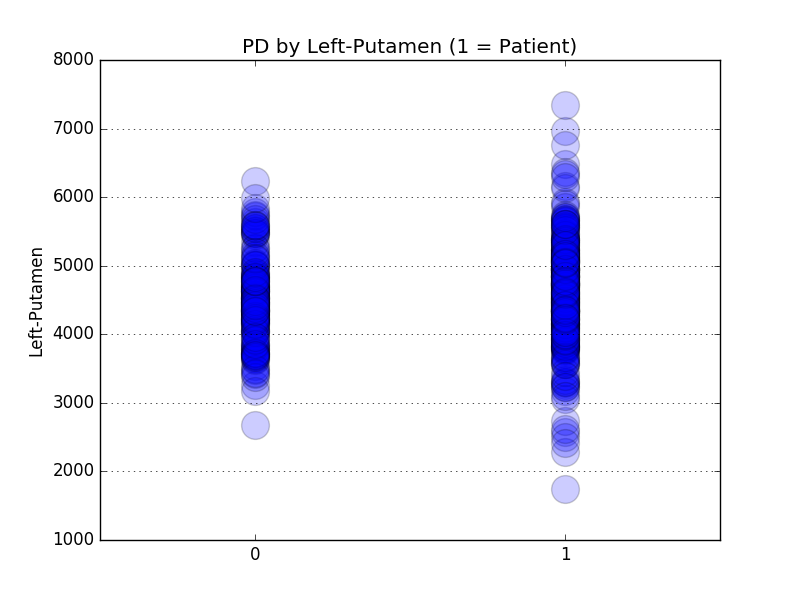}\label{fig:f2}}
  \\
    \subfloat[]{\includegraphics[width=0.5\textwidth]{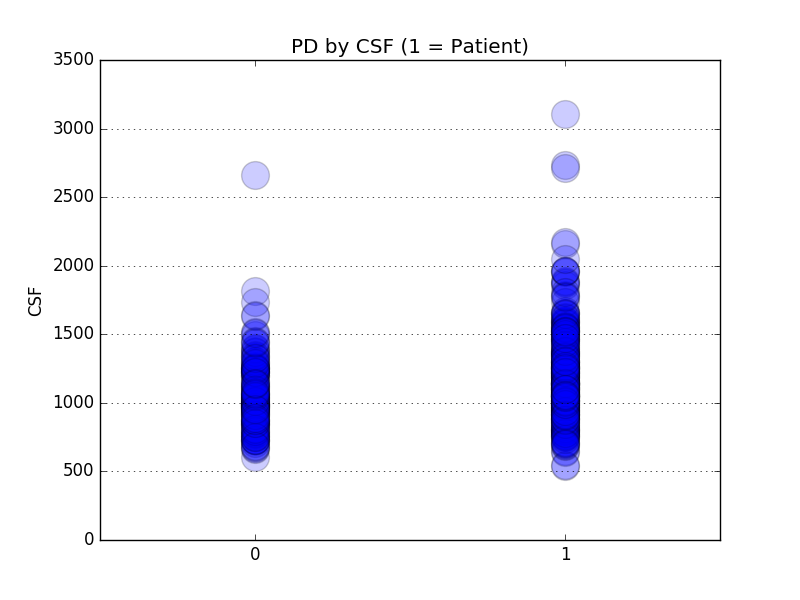}\label{fig:f1}}
  \hfill
  \subfloat[]{\includegraphics[width=0.5\textwidth]{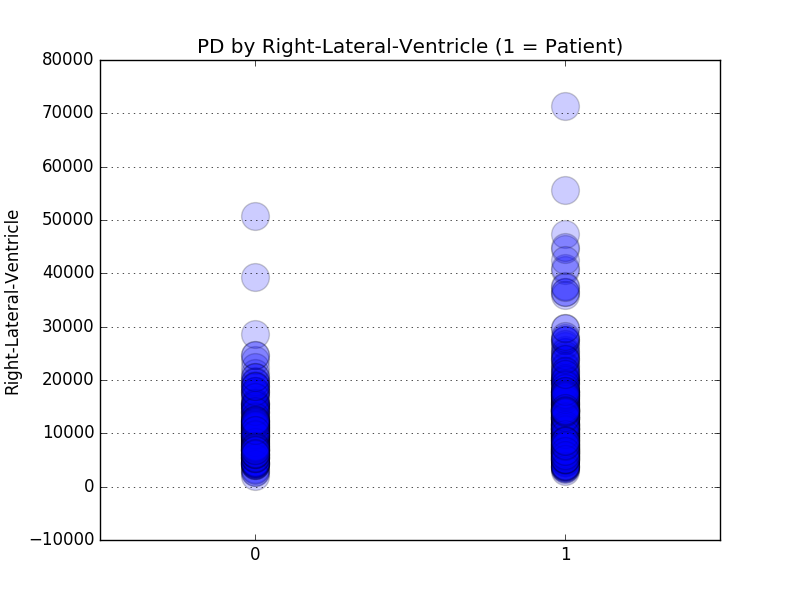}\label{fig:f2}}
  \caption{Data distributions in terms of the class labels and corresponding features, which are: (a) Left cerebellum volume. (b) Left putamen. (c) CSF. (d) Right lateral ventricle.}\setlength{\belowcaptionskip}{0pt}
  \label{fig:feateval}
\end{figure}

\begin{figure}[!tbp]
  \centering
  \subfloat[]{\includegraphics[width=0.5\textwidth]{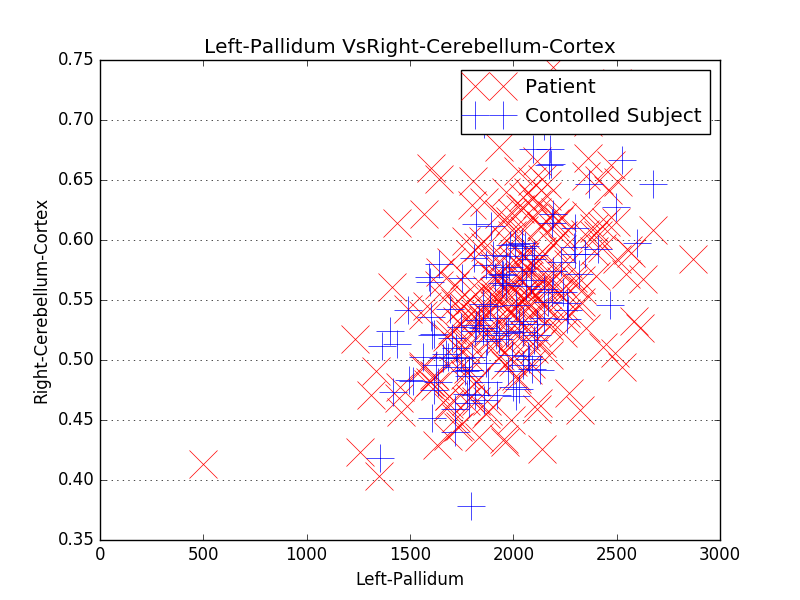}\label{fig:f1}}
  \hfill
  \subfloat[]{\includegraphics[width=0.5\textwidth]{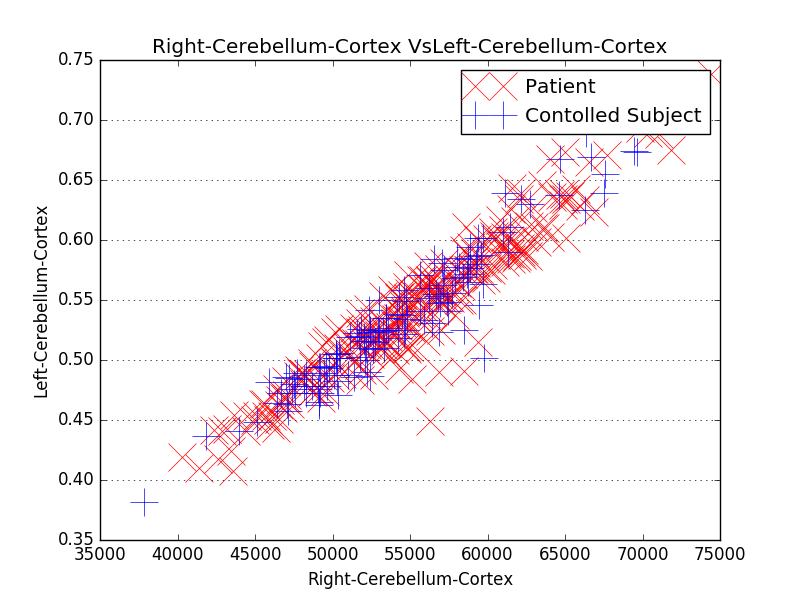}\label{fig:f2}}
  \caption{Data distribution based on the pair of features: (a) Left pallidum vs right cerebellum cortex. (b) Right cerebellum cortex vs left cerebellum cortex.}
  \label{fig:pairfeat}
\end{figure}

After preprocessing with FreeSurfer, a list of volume-based features is extracted for each subject. Also, age and sex are provided for the PPMI data on their website as of the patients' demographic information. Some evaluation has been done over the set of extracted features in terms of their discrimination ability. Since PD is an age related disease, the distribution of data in terms of age is plotted. Fig.\ref{fig:agelabel} shows the distribution of age in the dataset for the subjects with PD and HC labels.
The distribution of all the extracted features is plotted in terms of their ability to divide the data into two classes, PD and HC. Some of these distributions are shown in Fig.\ref{fig:feateval}. As can be seen in Fig.\ref{fig:feateval}(a), the subjects with PD have higher cerebellum cortex volume compared to the healthy ones. Furthermore, the distribution in Fig.\ref{fig:feateval}(b) and (c) illustrate that when people are in the PD category, their putamen and CSF volume size is intended to be enlarged. Fig.\ref{fig:feateval}(d) shows that the right lateral ventricle volume in PD is noticeably higher than in the normal subjects.
Another set of evaluations was performed over the extracted features. Data distribution for each pairs of features are plotted based on the corresponding class. Fig.\ref{fig:pairfeat} shows the distribution of data based on the two pairs of features including  Left pallidum vs right cerebellum cortex and right cerebellum cortex vs left cerebellum cortex. In both of them, two features tend to have bigger value when the subject is PD.

As explained in the previous section, three types of classifiers are used in this study. These algorithm are run over $507$ samples with $141$ features. The number of PD and control samples in this set of subjects are $340$ for PD and $167$ for HC. Since there is not enough balance for the data, we did data augmentation to b balance it. Since, the number of HC (negative) samples is not enough, we increase these samples just by creating a new set of negative samples calculated by subtracting the mean value from the current negative feature values. After doing data augmentation, the total number of samples is $673$ with $341$ PD (positive samples) and $332$ HC (negative samples).
Internal and external cross validation is applied with $K=10$ for external and $k=5$ for internal (parameter tunning cross validation). The number of selected samples for the training part is $536$ and for the test part, $67$. The number of PD and HC in each group is presented in Table \ref{tab:trtedata}.
\begin{table}[h]\small
\centering
\caption{Data balance in training and testing parts.}\setlength{\abovecaptionskip}{0pt}
\begin{tabular}{|c|c|c|c|}
\hline
         & PD  & Hc  & Total \\ \hline
Training & 307 & 229 & 536   \\ \hline
Test     & 34  & 33  & 67    \\ \hline
\end{tabular}
\label{tab:trtedata}
\end{table}

As mentioned before, the classification algorithm needs a set of parameters for tunning which is selected as follows:
\begin{itemize}
\item logistic Regression (LR):\\
Regularization = $[1e-1, 1e-2, 1e-3, 1e-4, 1e-5]$, Tolerance = [1e-1, 1e-2, 1e-3, 1e-4, 1e-5]
\item Random Forest (RF):\\
Number of estimator =$[5, 10, 15, 20, 25]$, Max depth = $[2-10]$
\item Support Vector Machine (SVM): \\
C = $[0.1, 1, 10, 100, 1000]$, Gamma = $[10, 1, 1e-1, 1e-2, 1e-3, 1e-4]$, kernels = $[linear, rbf, poly]$
\end{itemize}
The evaluation metrics used in this paper for comparing the results of the classification algorithms include accuracy for training and testing data and AUC (area under ROC curve).
Table \ref{tab:comp} shows the general comparison between these methods which is achieved by averaging the accuracy over 10-fold cross validation.  In the table there are two sets of results related to using age/sex feature or the classification built only on the extracted volume based features from FreeSurfer. As you can see, the best result is for RF either with age/sex feature or without it. Although the LR result is close to that. However, if we compare the results based on the training accuracy showing the ability of the classifier to learn a feature from the data, SVM-linear is the best one.

Based on the literature review, most studies use SPM with VBM toolbox for data analysis and MRI data feature extraction not only for PD evaluation, but also for other neuro diseases. In this paper, one of the important goals was to evaluate FreeSurfer in terms of preprocessing and feature extraction over T1 MR Images for PD subjects using machine learning techniques.  Generally, the experimental results show that the classification models need more information about the data that should be added to the current features as, these are low-level features and we need a set of high-level features as well. In future research, we are going to determine the useful general features that can be combined with the volume based features extracted from the PPMI data.
\begin{table*}[h]\small
	\centering
	\caption{Comparing performance of different classifiers}
	\label{tab:comp}
	\begin{filecontents*}{compmet.csv}
Methods/Criteria,Age/Sex Feature,Train Accuracy,Test Accuracy,AUC
RF,No,0.7667,0.7419,0.7398
RF,Yes,0.7944,0.7576,0.7545
LR,No,0.7673,0.7373,0.7319
LR,Yes,0.7705,0.7502,0.7479
NN,No,0.6757,0.6665,0.6642
NN,Yes,0.7006,0.6825,0.6791
SVM with rbf kernel,No,0.5044,0.5044,0.5
SVM with rbf kernel,Yes,0.5044,0.5044,0.5
SVM with linear kernel,No,0.7698,0.7091,0.7086
SVM with linear kernel,Yes,0.8015,0.7146,0.7138
\end{filecontents*}
\csvautotabular{compmet.csv}
\end{table*}
\section{Conclusion}
We presented an automatic MRI based CAD system for diagnosing Parkinson\textsc{\char13}s Disease (PD), the second common neuro degenerative disease affecting elderly people. This disease is exposed by the loss of neuro-transmitters that control body movements. Currently, there is no cure other than earlier diagnosis with better and more efficient treatment for patients. We used MR T1 images from the public PPMI PD dataset and FreeSurfer for feature extraction and preprocessing. The decision model for classification of the extracted feature data is based on LR, RF, and SVM methods. In the experimental results, we compare the ability of these three types of classifiers to diagnose PD. The results show that using MRI only has a potential for diagnosing PD. This approach will avoid exposing the brain to harmful radiation based scans. In future work, the efficiency of the proposed method could be improved by adding high level features to the current ones. In addition, the classification rate with MRI needs to be improved to get close to rate achieved by those using raditation based scanning.



\end{document}